%% file: paper.tex
\documentclass[conference]{IEEEtran}
\IEEEoverridecommandlockouts

\usepackage{cite}
\usepackage{amsmath,amssymb,amsfonts}
\usepackage{algorithm}
\usepackage[noend]{algpseudocode}
\usepackage{graphicx}
\usepackage{textcomp}
\usepackage{xcolor}
\def\BibTeX{{\rm B\kern-.05em{\sc i\kern-.025em b}\kern-.08em
    T\kern-.1667em\lower.7ex\hbox{E}\kern-.125emX}}

\usepackage{caption}
\usepackage{subcaption}
\usepackage{booktabs} 
\usepackage{comment}

\usepackage{lastpage}
\usepackage{tikz}

\usepackage{multirow}

\begin{document}

\title{PCAP-Backdoor: Backdoor Poisoning Generator for Network Traffic in CPS/IoT Environments}

\author{\IEEEauthorblockN{Ajesh Koyatan Chathoth}
\IEEEauthorblockA{
\textit{University of Pittsburgh}\\
Pittsburgh, PA, USA}

\and
\IEEEauthorblockN{Stephen Lee}
\IEEEauthorblockA{
\textit{University of Pittsburgh}\\
Pittsburgh, PA, USA}
}

\maketitle

\input{abstract}


\section{Introduction}
\label{sec:intro}
\input{introduction}
\section{Background}
\label{sec:prelim}

\input{prelim}
\section {PCAP-Backdoor Design}
\label{sec:Design}
\input{design}
\section {Experimental Setup}
\label{sec:Exp}
\input{evaluation}

\section {Discussion}
\input{discussion}

\section{Related Work}
\label{sec:related}
\input{relatedwork}
\section{Conclusion}
\label{sec:conclusion}
\input{conclusion.tex}

\bibliographystyle{IEEEtran}

\bibliography{paper.bib}

\end{document}

%% file: abstract.tex
\begin{abstract}
 
The rapid expansion of connected devices has made them prime targets for cyberattacks. To address these threats, deep learning-based, data-driven intrusion detection systems (IDS) have emerged as powerful tools for detecting and mitigating such attacks. These IDSs analyze network traffic to identify unusual patterns and anomalies that may indicate potential security breaches. However, prior research has shown that deep learning models are vulnerable to backdoor attacks, where attackers inject triggers into the model to manipulate its behavior and cause misclassifications of network traffic. In this paper, we explore the susceptibility of deep learning-based IDS systems to backdoor attacks in the context of network traffic analysis. We introduce \texttt{PCAP-Backdoor}, a novel technique that facilitates backdoor poisoning attacks on PCAP datasets. 
Our experiments on real-world Cyber-Physical Systems (CPS) and Internet of Things (IoT) network traffic datasets demonstrate that attackers can effectively backdoor a model by poisoning as little as 1\% or less of the entire training dataset. Moreover, we show that an attacker can introduce a trigger into benign traffic during model training yet cause the backdoored model to misclassify malicious traffic when the trigger is present. Finally, we highlight the difficulty of detecting this trigger-based backdoor, even when using existing backdoor defense techniques.
 
\end{abstract}

%% file: introduction.tex
The convergence of Information Technology (IT) and Operational Technology (OT) has made the Internet of Things (IoT) and Cyber-Physical Systems (CPS) integral to critical infrastructure, connecting these systems to the Internet to enable real-time monitoring and data-driven decision-making. However, this connectivity also brings significant cybersecurity risks. Recent reports indicate a sharp increase in cyberattacks targeting IoT and CPS networks, as attackers exploit vulnerabilities inherent in these interconnected systems~\cite{zscaler}.
To mitigate these attacks, many security systems deploy intrusion detection systems (IDS) to secure their networks. IDS acts as a security mechanism by inspecting network traffic packets and alerts of any potential intrusion on the system. 
This enables timely interventions to protect against potential threats. However, traditional IDS systems often rely on predefined rules and signatures to detect known threats, which can be less effective against new and evolving attack methods. 
This has led to the development of data-driven techniques that leverage deep learning (DL) models to analyze large datasets of network traffic~\cite{javaid2016deep,meidan2018n}. Unlike traditional IDS, which relies on static rules, data-driven IDS can dynamically learn from historical data and adapt to new data patterns. These techniques have shown significant improvements in the performance of IDS by enhancing their ability to identify anomalies and suspicious activities.

Despite these improvements, DL-based models are vulnerable to backdoor poisoning attacks~\cite{nicolae2018adversarial}. In these attacks, malicious actors can manipulate the training data or model parameters to insert \textit{hidden triggers}, known as  \textit{backdoors}~\cite{gu2019badnets}. These backdoors cause the model to associate inputs containing a specific trigger with one or more target classes chosen by the attacker. Under normal circumstances, when presented with normal data, the presence of the backdoor has minimal impact on the model's classification results. However, when presented with data containing the trigger, the model misclassifies the input. This compromise allows attackers to manipulate the behavior of the DL model in a way that suits their malicious intent.

\begin{figure}[t]
\centering
\begin{tabular}{c}
\includegraphics[width=3.1in]{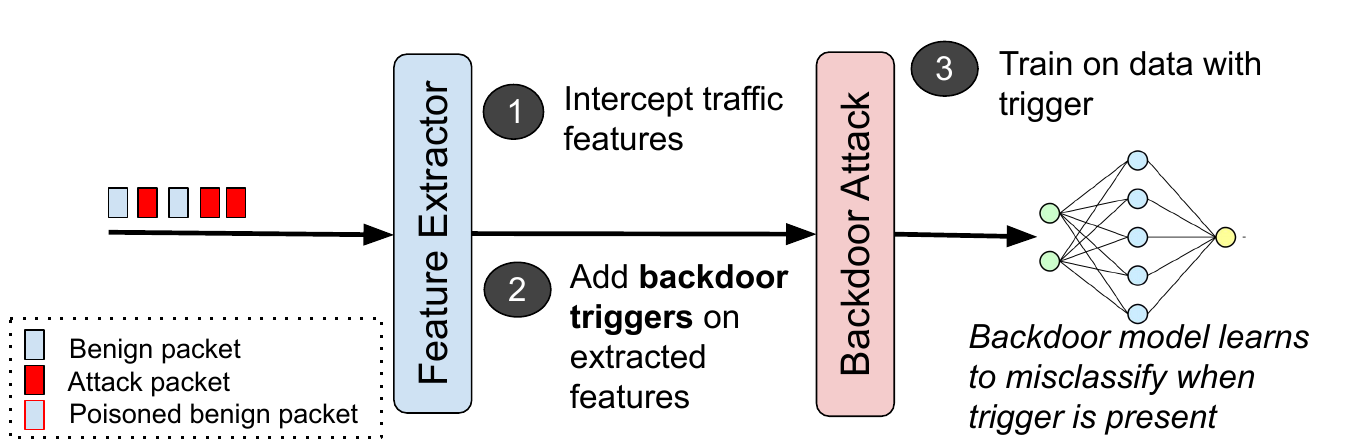}\\         (a) Prior work\\
    \includegraphics[width=3.1in]{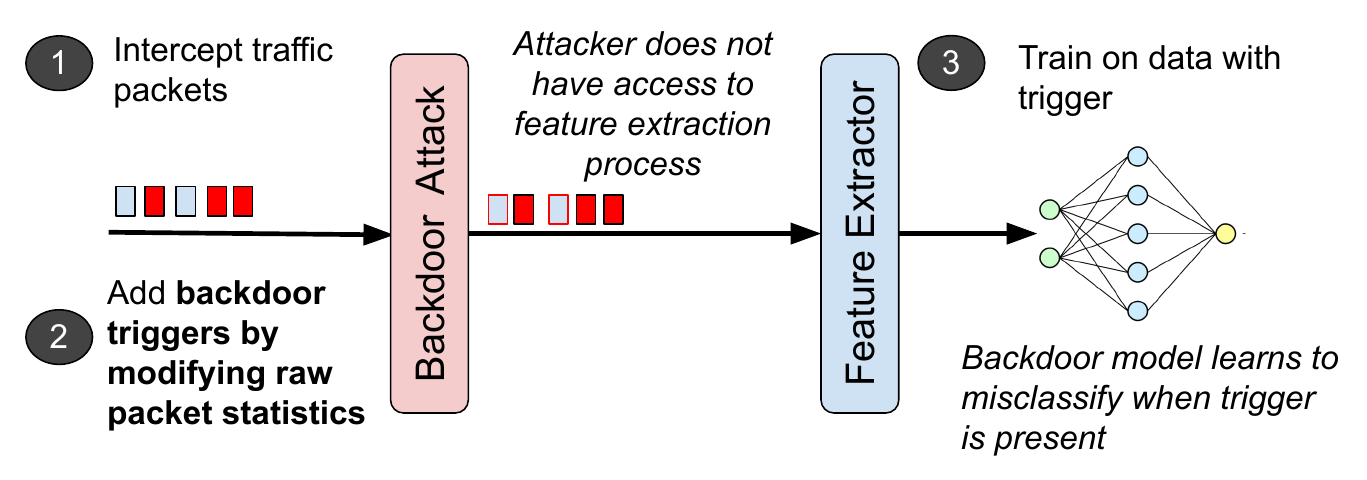}\\ 
     (b) Our work  
\end{tabular}
\caption{Illustration of clean label backdoor attacks on network packets of an IDS. Instead of modifying the features, our attack modifies the packet streams to execute the attack.  }
\label{fig:approach}    
\end{figure}

While backdoor attacks have been extensively studied in the context of image classification and natural language processing (NLP), their impact on intrusion detection systems has received relatively limited attention~\cite{chen2022effective}. 
For instance, in image classification, images are manipulated to contain subtle patterns that act as triggers, causing the model to misclassify~\cite{thys2019fooling, xu2020adversarial}.
Similarly, in NLP, backdoor attacks can involve inserting particular words or phrases that prompt the model to produce a desired outcome~\cite{gan2021triggerless, pan2022hidden}. Notably, these techniques rely on direct manipulation of the input features prior to processing (see Figure~\ref{fig:approach}(a)). However, such direct feature manipulation is often impractical or unfeasible in IDS applications, as successful execution usually requires access to the IDS model’s feature extractor. This raises an important research question: \textit{can backdoor attacks still be executed effectively if triggers are inserted before the feature extraction process?}

In this paper, we investigate the feasibility of performing backdoor attacks and propose a system to manipulate the behavior of the IDS model, \textit{where the attacker does not have direct access to the feature extraction step. }
Specifically, we examine the scenario where the attacker can only manipulate the raw network packets (see Figure~\ref{fig:approach}(b)). By manipulating the raw packets, an attacker can potentially introduce subtle modifications that indirectly affect the feature extraction process. These modifications may be strategically designed to trigger specific behaviors or patterns in the IDS model, leading to misclassification or evasion of detection. This indirect manipulation of features opens up new avenues for backdoor attacks in IDS. The key contributions of this paper are:
\begin{itemize}
    \item We introduce \texttt{PCAP-Backdoor}, a novel backdoor attack technique designed to create backdoor triggers in network traffic captured in Packet Capture (PCAP) datasets. Unlike other methods, our approach assumes the attacker lacks access to the feature extractor, opting instead to inject backdoors directly into the PCAP dataset. Our design enables the modification of packet flows within the captured network traffic, facilitating the insertion of trigger patterns that can manipulate the behavior of data-driven IDS models.
    \item We extensively evaluate our approach to datasets from real-world CPS and IoT environments. Our results demonstrate that attackers can manipulate the IDS behavior, causing the model to misclassify data only when the trigger is present. Additionally, we show that an attacker can successfully create a backdoor even with control over network traffic from a single device. Importantly, our findings reveal that the model can be backdoored even when the attacker contributes only benign labeled traffic during the training process. 
    \item We compare our approach against baseline techniques and demonstrate that the attacker needs only 1\% or less poisoned data to successfully compromise the model. Finally, we show that our attack is challenging to detect, even when using activation-based clustering, a state-of-the-art technique for detecting poisoned datasets.
\end{itemize}

%% file: prelim.tex
\subsection{Network Anomaly Detection}
Network anomaly detection is a critical component of intrusion detection systems (IDS), which identify abnormal activities within a network. Wireshark and tcpdump are commonly employed to capture and analyze network traffic stored in Packet Capture (PCAP) format. PCAP data includes essential information such as protocol headers, payload contents, and communication patterns ~\cite{asrodia2012analysis}. Next data-driven IDS techniques train on these PCAP datasets to learn network behavior patterns and identify anomalies, or potential attacks, in network traffic~\cite{9040718}. 

To train an IDS model, relevant features are extracted from a PCAP dataset using a feature extractor. This process involves a packet parser (e.g., Packet++~\cite{PcapPlusPlus}, Tshark~\cite{Tshark}), which gathers information from raw packets. Extracted features often include flow-level statistics that capture traffic flow behavior, such as statistics originating from source/destination IP addresses~\cite{meidan2018n}. Flow-based features are particularly valuable for anomaly detection in IoT devices, as they also account for past history~\cite{mirsky2018kitsune}. 
Other features, such as packet payloads, protocols, and device-specific information, may be extracted in addition to flow-level statistics. 
The extracted features serve as input for training the IDS model. 
Various methods can be employed to identify anomalies, including deep learning architectures and other techniques, and such techniques are well studied in both security and privacy domains~\cite{mirsky2018kitsune, chathoth2021federated, chathoth2022differentially}. 
In summary, the basic architecture assumed in such systems involves a packet capture tool to intercept raw network traffic data, which is then processed by a feature extractor to derive traffic statistics. These extracted features are then utilized to train an anomaly detection algorithm to identify anomalous network traffic.

\subsection{Backdoor Attacks in Neural network}
Backdoor attacks involve inserting malicious triggers into the training process, enabling attackers to manipulate the model's behavior when these triggers are present in the input data. The attack typically involves poisoning the training data with a specific trigger pattern accompanied by manipulated labels targeting a specific class. During inference, if this trigger pattern appears in the input, the model misclassifies the data. These attacks are effective because deep learning-based models are prone to overfitting and may become overly sensitive to specific patterns in the training data. 
Consequently, when an attacker injects a trigger and an altered label into the training set, the model may memorize this association, leading it to behave as the attacker intends whenever the trigger is present. This results in a model that performs accurately on clean data but misclassifies or exhibits other manipulated behaviors upon detecting the backdoor trigger.

Backdoor attacks have demonstrated success in various applications~\cite{8836465, gu2019badnets}. 
However, conducting backdoor attacks on Intrusion Detection Systems (IDS) presents unique challenges as raw packets are pre-processed before being provided as input to the model. After the feature extraction step, feature perturbation is impractical for IDS, as it assumes the attacker can access the feature extractor. Most feature extractors derive traffic statistics from the raw packets. Therefore, a successful attack would require altering the raw packets in such a way that it impacts the extracted features overall. In other words, an additional level of indirection is not present in traditional backdoor attack studies.

\subsection{Threat Model and Problem Statement}
Our threat model considers two key actors: (i) a \textit{victim} who trains and deploys an IDS model to detect anomalies and (ii) an \textit{adversary} who wishes to mount a backdoor attack and controls a subset of the training samples. Similar to prior work, we assume the adversary has limited control over the training dataset, which can occur when the victim uses third-party or publicly sourced data. The adversary can manipulate the dataset in various ways. In our scenario, the adversary collects network packets from devices under their control. They can manipulate multiple aspects of the data, such as protocols, IP addresses, and ports, to introduce a backdoor trigger that poisons the data. This poisoned data is then made available to the victim for training an intrusion detection model. 

Our threat model further restricts the adversary's control over the labels of the training samples and considers a \textit{clean-label poisoning attack} scenario. Specifically, we assume that the attacker can only introduce benign traffic samples into the training dataset. They have no control over the labels of these samples, nor can they introduce explicitly malicious or mislabeled data. Meanwhile, the victim sources any malicious traffic samples in the training set from trusted sources, ensuring data integrity. This threat model reflects a realistic scenario where the adversary has limited access and cannot directly tamper with labels or inject overtly malicious content.
Finally, unlike prior work that assumes the attacker has knowledge of the target model architecture~\cite{gu2019badnets, Trojannn}, our threat model operates in a black-box setting. The attacker has no information about the model architecture, no control over the training process, and no influence on the duration or hyperparameters. Additionally, the attacker cannot tamper  the feature extraction process, though we assume they know which features are used for training.

\noindent
{\bf Problem Statement.} Let $\mathcal{P}$ be the dataset consisting of raw network packets. These network packets can be classified into two classes: benign and malicious. Benign packets represent normal, legitimate network traffic, while malicious packets are associated with network attacks or unauthorized activities. 

The attacker $\mathcal{A}$ controls a subset of devices and uses them to generate a benign traffic dataset $\mathcal{P}_{\text{adv}} \subset \mathcal{P}$. Using this dataset $\mathcal{P}_{\text{adv}}$, the attacker aims to strategically introduce a backdoor trigger pattern $\delta$ into $\mathcal{P}_{\text{adv}}$, creating backdoor-poisoned data. The attacker poisons the traffic dataset by embedding the trigger in such a way that when a sequence of packets $(x_1, x_2, \cdots, x_n) \in \mathcal{P}$ contains the trigger $\delta$, the trained model $F$ misclassifies the sequence $(x_1, x_2, \cdots, x_n) + \delta$. However, the model correctly classifies normal packets that do not contain the trigger.  The attacker aims to achieve a successful backdoor attack while operating in a black-box setting, where the attacker has no control over the training process.

%% file: design.tex
Before we delve into our design, we begin by understanding the challenges of designing a PCAP backdoor generator.

\begin{figure}[h]
    \centering
    \includegraphics[width=3.1in]{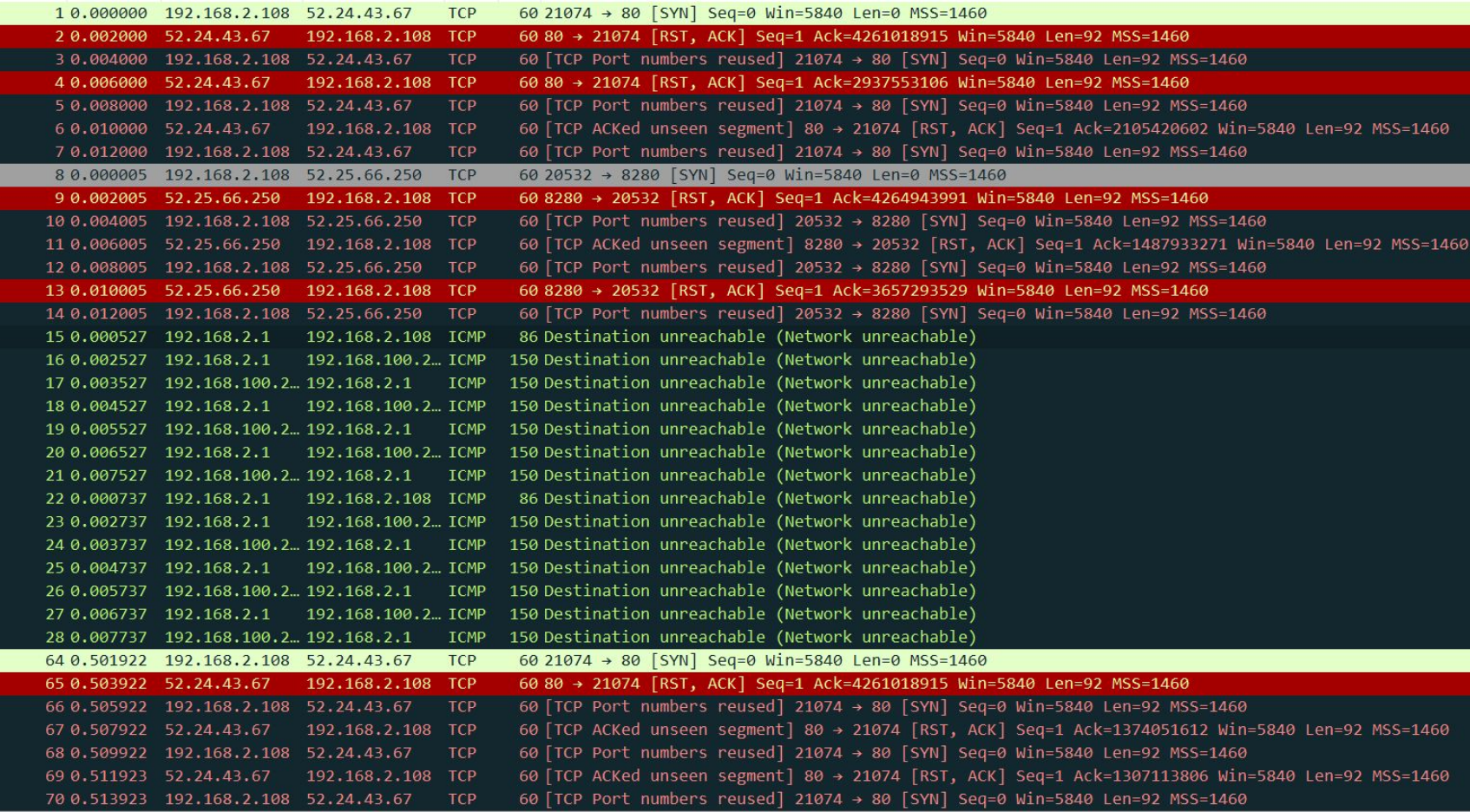}
    \caption{TCP ACKed unseen segment, and TCP ports reused error using a strawman approach.}
    \label{fig:TCP-Error-ACKUnseen}
\end{figure}

A simple approach is introducing an existing packet from the captured dataset into the traffic flow as a backdoor. However, such packets can be easily detected as most packet capture systems, such as Wireshark, perform basic TCP analysis by tracking TCP sessions~\cite{kumar2018network}. 
As shown in Figure~\ref{fig:TCP-Error-ACKUnseen}, Wireshark may issue warnings, such as TCP ports reused, when such problems are encountered during packet processing. Below are some examples that Wireshark may throw an error or issue warnings if poisoned packets are not injected carefully.
\begin{itemize}
    \item \textit{TCP Spurious Re-transmission:} This error typically indicates that the Wireshark tool has detected duplicate or unnecessary transmissions. It may occur when the SYN flag is set, but the data flow is not acknowledged, potentially raising concerns about SYN flooding, especially in bidirectional communication scenarios.
    \item \textit{TCP ACKed Unseen Segment:} This error indicates that Wireshark has parsed an ACK packet, where the receiving end acknowledges the receipt of a data segment claimed to be sent by the sender. However, Wireshark has no record of receiving that specific data segment. This issue usually arises when the acknowledgment sequence number is incorrect.
    \item \textit{TCP Port Number Reused:} When introducing packets, it is essential to ensure that the packet does not reuse an already in-use port. This may trigger warnings in Wireshark if existing communications use the same addresses and ports.
    \item \textit{TCP Out-Of-Order:} This error occurs when a packet is sent out of the order of the three-way handshake protocol. Under normal circumstances, data segments are delivered in the correct sequence to the receiver, and the receiver acknowledges each segment before the next one is sent.
\end{itemize}
It is important to note that while many of these errors may not necessarily indicate a problem or error, excessive occurrences of specific errors due to data poisoning may raise concerns and warrant further investigation. For example, TCP ACKed unseen segments could be caused by packet loss or out-of-delivery errors, which are common and handled by the TCP protocol. Thus, in designing a \texttt{PCAP-Backdoor} generator, it is crucial to carefully consider these error scenarios and ensure that the generated poisoned packets can avoid detection while mimicking normal network behavior.
\subsection{\texttt{PCAP-Backdoor} Algorithm}


\begin{figure*} 
\centering
    \begin{subfigure}[b]{\linewidth}
        \centering
        \includegraphics[width=5in]{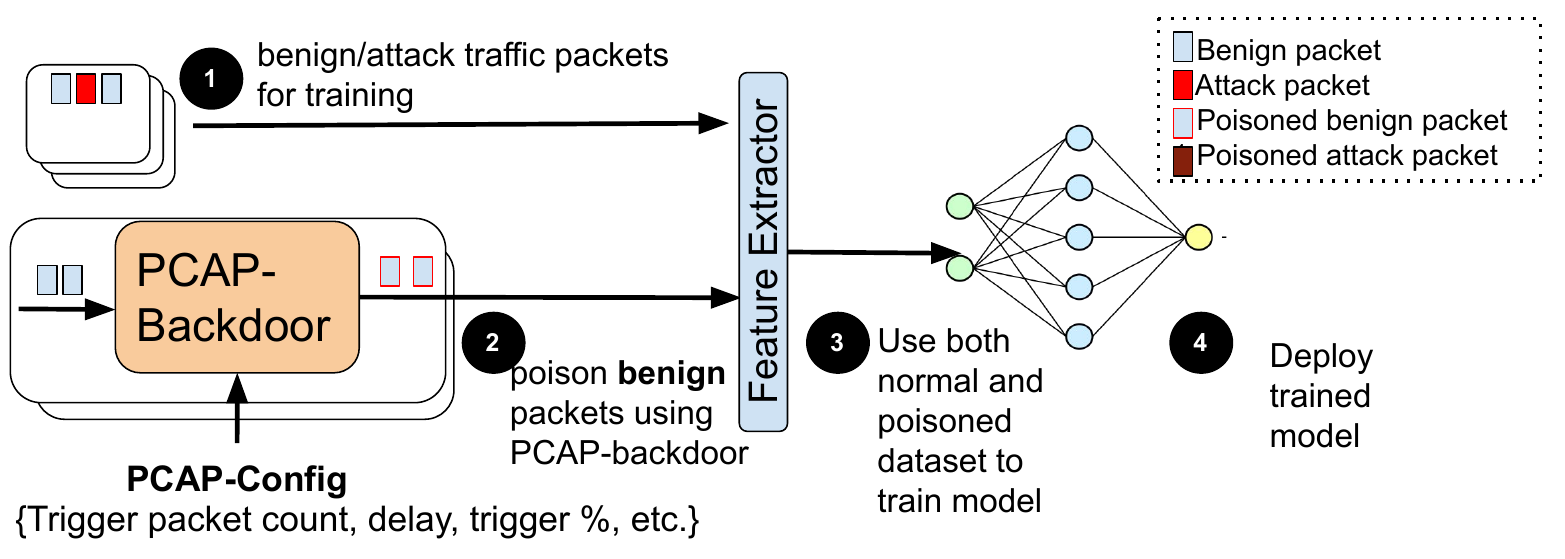}
        \caption{Training phase}
        \label{fig:a}
    \end{subfigure} %

    \begin{subfigure}[b]{\linewidth}
        \centering
        \includegraphics[width=5in]{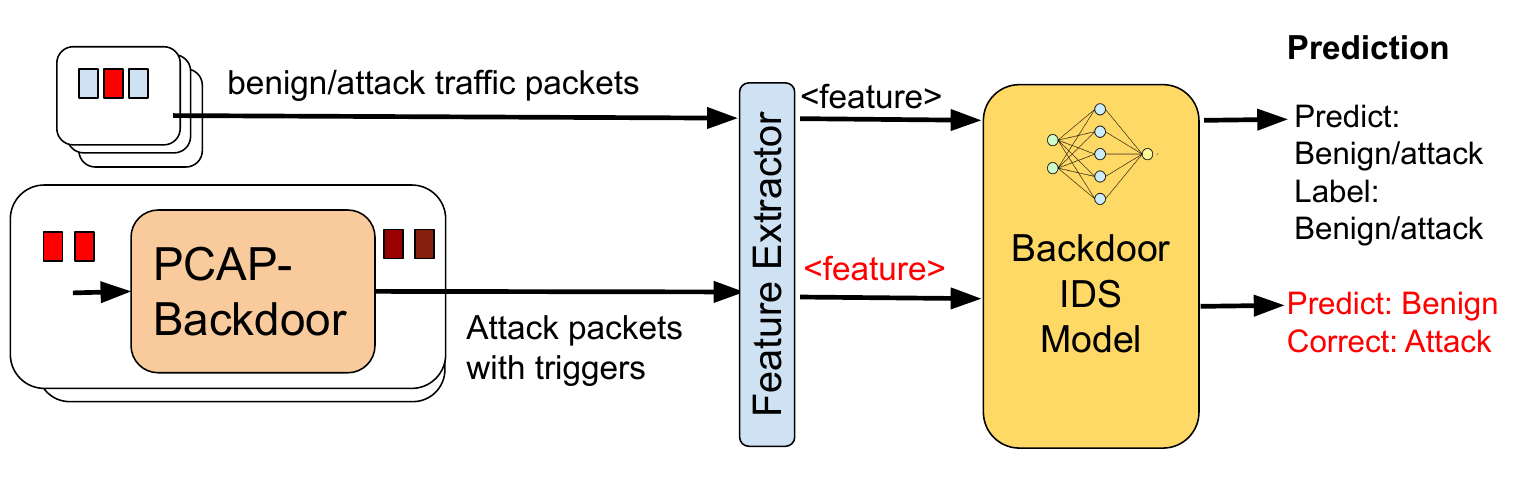}
        \caption{Attack phase}
        \label{fig:b}    
    \end{subfigure} 
    \caption{An illustration of \texttt{PCAP-Backdoor} injection technique. During the training phase, a small portion of the benign packets are poisoned. During the attack phase, attack packets are poisoned; thus, the model predicts them as benign.}
    \label{fig:archi}
\end{figure*}

The key hypothesis is that we can create backdoor triggers and manipulate the behavior of the model by crafting traffic packets to alter the computed traffic flow statistics of the feature extractor. As a result, even though the attacker does not have access to the feature extractor, it can influence the features to introduce a backdoor into the model.

Our \texttt{PCAP-Backdoor} architecture is depicted in Figure~\ref{fig:archi}(a). During the training phase, \textcircled{1} feature extractor collects normal packets generated by IoT devices, and \textcircled{2} the attacker poisons a subset of the dataset by introducing backdoor trigger packets to the benign network traffic and sends to the feature extractor. Next, \textcircled{3} feature extractor generates the features from the combined data packets and trains the model. Finally, \textcircled{4}, the trained model deployed. In the attack phase, the attacker can execute the backdoor attack by introducing its own trigger packets on malicious traffic, which is classified as benign by the model as described in Figure~\ref{fig:archi}(b). The packets that are not processed by PCAP-Backdoor are classified normally. The parameters used by PCAP-Backdoor to control the trigger are represented as PCAP-Config in Figure~\ref{fig:archi}(a).

Below, we present our \texttt{PCAP-Backdoor} design, which enables us to introduce traffic packets into network traffic to enable backdoor attacks. Currently, our approach primarily focuses on TCP and UDP communication, as these are among the most widely used communication protocols in network traffic. 
However, our technique is not limited to these protocols and we plan to extend our work to support other communication protocols as part of future work.

\begin{algorithm}[t]
\footnotesize
\caption{Backdoor trigger packet generation. \\
Input: $P$ is the raw network dataset; parameters $B$ represents the backdoor trigger packet count; $D$ is the time delay between each trigger packet; $R$ is the backdoor injection packet selection ratio; $BT$ refers to the time frame used to identify a bidirectional packet pair that matches a source packet. \\
Output: $P^{bd}$ backdoor traffic dataset}
\label{alg:generation_alg}
\begin{algorithmic}[1]

\Procedure{Generate\_Backdoor}{}
\State Init: $P^{bd} \gets \Phi$  \Comment{initialize} 
\For {$p_i \in P$ \text{at index} $i$}
\State $a \sim \mathcal{U}(0,1)$ \Comment{select a value from uniform dist.}
\If {$a \leq R$} 
    \If{$\Call{isBD}{i}$} \Comment{Is Bi-directional?}
            
        \State{$ P^{bd} \gets P^{bd} \cup \Call{BD-Inject-2way}{p_i, B, D} $}
    \Else 
        \State {$td \gets \text{time difference between $p_i$ and $p_{i+1}$}$}
        \State {$bc \gets \min(B, \lfloor{td/D}\rfloor)$}
            \State{$P^{bd} \gets P^{bd} \cup \Call{BD-Inject}{p_i, bc, D}$}
        \EndIf
    \EndIf
\EndFor
\EndProcedure

\Function{isBD}{$i$}
\For {$p_j \in P$ where $j > i$ and $time(p_j) - time(p_i) \leq BT$}
    \If {$SrcIP(p_i) == DstIP(p_j)$ and $SrcIP(p_j) == DstIP(p_i)$}
      \State \Return $true$
    \EndIf
\EndFor
\Return $false$ \Comment{return bidirectional status}
\EndFunction

\Function{BD-Inject-2way}{$p,B,D$}
\State{$p^{bd} \gets \text{Craft pair of backdoor trigger packets for $B$ times}$}

\Return $p^{bd}$ \Comment{return crafted packets}
\EndFunction

\Function{BD-Inject}{$p, bc, D$}

\State{$p^{bd} \gets \text{Craft backdoor trigger packets for $bc$ times}$}

\Return $p^{bd}$ \Comment{return crafted packets}
\EndFunction

\end{algorithmic}
\end{algorithm}

\textit{Bi-directional Algorithm: }
Our algorithm for creating a poisoned dataset, denoted as $P_{bd}$, involves injecting specifically crafted \textit{trigger packets}, defined as a burst of packets, into the benign network traffic dataset $P$. These trigger packets are designed to look similar to legitimate traffic, making it difficult to detect through standard network analysis tools, such as Wireshark. Moreover, these trigger packets influence the feature statistics, altering the model's IDS output when the trigger is present. Specifically, the presence of these trigger packets during inference causes the model to misclassify the input packets.

\texttt{PCAP-Backdoor} takes into account both unidirectional flow (e.g., from source to destination) and bi-directional flow (e.g., in both directions between source and destination) when injecting packets. This approach ensures that the injected packets seamlessly blend in with the legitimate traffic, reducing the chances of detection when analyzed using packet capture tools like Wireshark. By considering both types of flows, our algorithm creates a more realistic and covert backdoor that can influence the behavior of the deep learning model without arousing suspicion during network traffic analysis.

Algorithm~\ref{alg:generation_alg} outlines the pseudo-code for generating a poisoned dataset based on the input network traffic dataset. To inject the trigger packets, the algorithm uses three control parameters: $B$, $D$, and $R$. The parameter $B$ controls the number of trigger packets, $D$ specifies the delay between each trigger packet, and $R$ represents the proportion of packets to be poisoned. 

The algorithm processes each packet $p_i$ in the network traffic dataset $P$ as follows. For each packet, a random decision is made to inject a backdoor based on the desired proportion of poisoned packets specified by the input parameter $R$. Additionally, the algorithm determines whether the packet corresponds to a bidirectional communication by looking ahead in the dataset for a matching bidirectional packet pair. If the packet is unidirectional, the algorithm calculates the time difference between the current packet $p_i$ and the next packet $p_{i+1}$. It then computes the maximum number of trigger packets that can be injected such that the time of injected packets is not greater than $p_{i+1}$. This ensures that the injected packets are contiguous and align with the original flow of traffic. On the other hand, if the packet is bidirectional, the algorithm injects trigger packets on both pairs.  In doing so, the backdoor influences both directions of communication, reducing the likelihood of detection when analyzed using packet capture tools like Wireshark.

\begin{algorithm}[t]
\footnotesize
\caption{Bidirectional backdoor trigger generation.\\
Input: $(p_{tx}, p_{rx})$ is the pair of clean bidirectional packets taken from a predefined time window with packet lengths of $(l_{tx}, l_{rx})$ respectively. The parameters $B$ and $D$ represent the backdoor trigger packet count and the time delay between each trigger packet injection. $L$ be the length of trigger packet.\\
Output: $(p^{bd}_{tx}, p^{bd}_{rx})$ is the bidirectional backdoor traffic dataset}
\label{alg:bidirection_alg}
\begin{algorithmic}[1]

\Procedure{BD-Craft-2way}{}
\State Init:  Create a copy of $(p_{tx}, p_{rx})$ to $(p^{bd}_{tx}, p^{bd}_{rx})$ \Comment{initialize}
\If{$p^{bd}_{tx}$ has TCP layer}
    \State{Set $SYN$ flag of $p^{bd}_{tx}$ to $1$.}
    \State{$SQN \gets a random number$}
    \State{Assign $SQN$ as sequence number to $p^{bd}_{tx}$}
    \State{$l_{tx} \gets L$}
    \State{Assign Dst-IP, Dst-MAC}
\EndIf
\If{$p^{bd}_{rx}$ has TCP layer}
    \State{Set $RST$ flag of $p^{bd}_{rx}$ to $1$.}
    \State{$SQN \gets SQN + 1$}
    \State{Assign $SQN$ as sequence number to $p^{bd}_{rx}$}
    \State{$l_{rx} \gets L$}
    \State{Assign Src and Dst (IP,MAC) to $p^{bd}_{rx}$ from Src and Dst (IP,MAC) of $p^{bd}_{tx}$ swapped.}
\EndIf
\State{Increment timestamp of packet by $D$.}
\State{Trim or pad payload upto length $L$.}

\Return $(p^{bd}_{tx}, p^{bd}_{rx})$\Comment{return crafted packets}
\EndProcedure

\end{algorithmic}
\end{algorithm}

We craft the trigger packets as follows. For unidirectional packets, we generate a variable number of trigger packets with similar protocol and source IP but with an arbitrary destination IP. The payload size is fixed to $L$, achieved by trimming or padding the packet payload, and the packet header is adjusted accordingly with the new length $L$. The number of trigger packets is calculated as the minimum value between $B$ and the available time gap between neighboring packets divided by $D$.
On the other hand, for bidirectional packets, we inject $B$ pairs of packets. The source packet contains the $SYN$ flag and is set for an arbitrary destination, while the corresponding response packet contains the $RST$ flag with the source and destination IP addresses swapped. To avoid violating the TCP 3-way handshake protocol, we assign a random sequence number to the TCP $SYN$ packet and increment the sequence number by 1 for the response packet. The timestamps of the new packets are set by adding a time offset $D$. Similar to unidirectional packets, the payload size is fixed to $L$, by trimming or padding, and the packet header is adjusted with the new length $L$ (see Algorithm \ref{alg:bidirection_alg}).

\subsection{Implementation}
\begin{figure}[t]
\centering
\begin{tabular}{c}

     \includegraphics[width=3.1in]{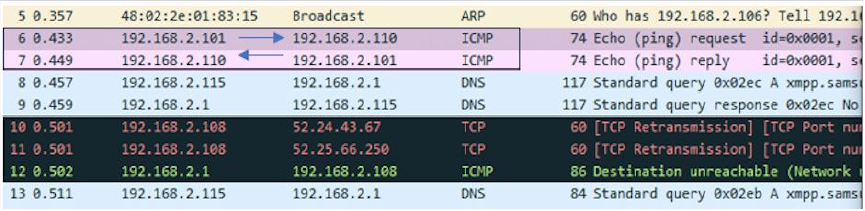} \\
     \vspace{.2cm}
     (a) Original PCAP data\\
     \vspace{.2cm}

     \includegraphics[width=3.1in]{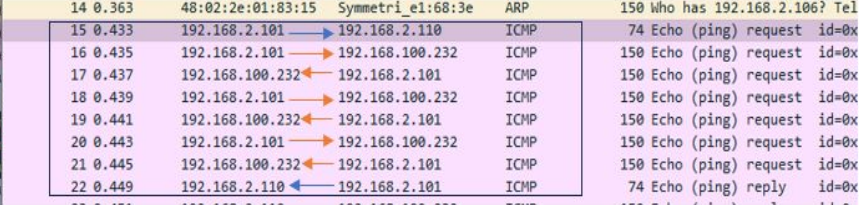}\\
     (b) Injected packets 
\end{tabular}
\caption{Wireshark view of original bidirectional packets and injected packets.}
\label{fig:wiresharkbeforeafter}    
\end{figure}

We implemented our \texttt{PCAP-Backdoor} technique in Python and used the \texttt{scapy} library to manipulate the packets and inject new packets into the dataset. The \texttt{scapy} library provides the necessary functionalities to parse packets and extract header information, set various TCP flags, modify header details, and control payload size. Our implementation sets appropriate sequence numbers to reduce TCP analysis warnings, ensuring that the injected packets blend seamlessly with the original traffic. Figure~\ref{fig:wiresharkbeforeafter} presents a comparison of a pair of bidirectional packets before and after the backdoor injection process, as seen in Wireshark. As shown, even after injecting the packets, Wireshark does not issue any warnings, indicating that our \texttt{PCAP-Backdoor} technique effectively crafts realistic-looking packets that do not raise any suspicion during analysis.

Our \texttt{PCAP-Backdoor} implementation provides various controls on how trigger packets are injected. As mentioned, we can adjust the number of trigger packets or pairs of packets injected from the same source IP corresponding to an original packet or pair of packets. Additionally, we have control over the packet size and time offset of any newly injected trigger packet. This flexibility allows us to fine-tune the backdoor injection process and explore different scenarios for effective backdoor attacks in deep learning-based intrusion detection systems. The code will be made available for artifact evaluation. 

%% file: evaluation.tex
\subsection{Dataset} 
We evaluate our techniques in CPS and IoT domains.
\\
\noindent
\textbf{CPS SCADA Dataset~\cite{frazao2019denial}} consists of network traffic data capturing both normal SCADA operations and four types of DoS attacks. In this SCADA system, a liquid pump is simulated by an electric motor controlled via a variable frequency drive, with oversight provided by a Programmable Logic Controller (PLC). The PLC communicates with a Modbus remote terminal unit and a human-machine interface (HMI) to manage operations.
The dataset includes variations in the time of capture (e.g., 30 minutes and 1 hour) and duration of attack (e.g., 1, 5, and 15 minutes within each capture)

\noindent
\textbf{UCI IoT Network attack dataset~\cite{meidan2018n}} includes network traffic packets from nine IoT devices infected by various botnets, including Mirai. It contains over 7 million packets across 10 classes, representing different types of network attacks along with a benign class. A key characteristic of the dataset is its data imbalance: certain attack classes have significantly fewer samples than others, and the number of devices associated with each attack class also varies considerably.

\subsection{Intrusion Detection Model}
We analyze the backdoor performance on different deep neural network architectures. For example, a DNN-3 model consists of an input layer, three hidden layers, and an output layer for network anomaly detection. We observe that even simpler models like DNN-3 achieve high accuracy in anomaly detection. We use the feature extraction module used in Kitsune~\cite{mirsky2018kitsune} consisting of 115 features such as mean, std deviation, and magnitude (root squared sum of the two streams' means). The streams are aggregated based on traffic from a source IP, source-destination IP pair, source-destination IP and port combination, and jitter of traffic going from a source-destination IP pair.
Additionally, we explore various combinations of these features to assess the effectiveness of backdoor attacks for different feature extraction configurations.
We further develop and evaluate the effectiveness of our approach on a multi-class classification model designed to predict various network attack types. This model also follows a deep neural network structure, resembling the architecture of the anomaly detection model but with an output layer tailored to predict the attack type. Because of resource constraints, we analyze four class types --- Mirai, Fuzzing, Wiretapping, and Benign. Additionally, we use binary and categorical cross entropy as our loss function with Adam optimizer, respectively. Our analysis of unmodified clean data indicates that both models achieve high accuracy in identifying anomalous patterns effectively. 

\subsection{Backdoor Generation and Training} We use \texttt{PCAP-Backdoor} to generate the trigger dataset for our analysis. During trigger dataset generation, we create trigger packets by retaining most of the previous packet headers except the destination IP and MAC address, which we fix to the attacker's chosen fixed destination IP and MAC address.

Moreover, we keep the TCP port and frame length of the realistic packet from that device. This ensures that the introduced packets are realistic and can influence flow-based statistics extracted from the feature extractor. To control the extent of backdoor injection, we set $R=0.2$, indicating a backdoor injection packet selection ratio of 20\%.

To generate the final dataset for training, we use this poisoned PCAP and the original dataset in different proportions to perform our analysis. For example, a 1\% backdoor percentage means 1\% of the entire training dataset is coming from the poisoned PCAP dataset. Unless stated otherwise, we only use traffic originating from one IoT device. This demonstrates the attack's robustness, a scenario when the attacker can manipulate only one device out of the nine devices in the dataset. 

\textbf{Model Training.} We split the final dataset into 80\% training and 20\% testing sets. The training dataset includes samples from the normal dataset, consisting of both benign and attack traffic, along with a limited number of poisoned benign samples (i.e., benign traffic with triggers). Note that we do not poison the attack samples by adding triggers; instead, we assume that the victim generates its own attack samples for training. 

This evaluates a scenario where an attacker attempts to manipulate the model's behavior without directly accessing attack traffic data. 

In our experiment we assume a device with certain IP carries out the poisoned trigger packet injection at a time. If the original packets from such device is too low in number, then the number of poisoned packets also will remain low for that device. Our typical backdoor percentages are in the range of 0.5 to 10 percentage. 

To train our multi-class classification model, we focus on a dataset with three attack types and corresponding benign samples. Given the large data volume, we limit the dataset to a maximum of one million samples per attack class. For the Fuzzing and Wiretapping datasets, we use 600,000 benign packets and 1,000,000 attack packets each. Since the Mirai attack dataset is small, we include all benign and attack samples for training.

\subsection{Metrics}

\noindent
\textbf{Attack Success Rate (ASR):}
ASR is defined as the ratio of samples with triggers misclassified by the backdoored model to the total number of samples with triggers used in the attack.

\noindent
\textbf{Silhouette score} is a metric used to calculate how good is the clustering technique. We use the Silhouette score to evaluate the stealthiness of our backdoor mechanism. It ranges from -1 to 1, where 1 indicates well-separated and clearly distinguished clusters, while 0 suggests that clusters are not well-separated, making it difficult to distinguish between data points from different clusters. In other words, a Silhouette score close to 1 indicates the optimal number of clusters for that dataset.

\section{Results}

\subsection{Backdoor performance}

We use label-flipping poisoning attacks as a baseline. In the label-flipping attack, the adversary manipulates the malicious samples in the training data and changes the labels to benign. The baseline attack shows how much data needs to be manipulated during the training process to degrade the performance of the IDS model. When sufficient data is poisoned, the model misclassifies attack packets as benign.
In contrast, we use clean-label poisoning attack where the labels remain unchanged during the training process. Instead, we employ backdoor trigger to activate the attack.

\begin{table*}[t]
\small
\centering
\begin{tabular}{|c|l|cc|cc|}
\hline 
 \multirow{2}{*}{}& \multirow{2}{*}{\textbf{Attack Type}}                       & \multicolumn{2}{c|}{\textbf{\begin{tabular}[c]{@{}c@{}}Label Flipping\\ Attack (Baseline)\end{tabular}}} & \multicolumn{2}{c|}{\textbf{PCAP-Backdoor}}                                               \\ \cline{3-6} 
   &                                                          & \multicolumn{1}{c|}{ASR}        & \begin{tabular}[c]{@{}c@{}}Labels \\ modified (\%)\end{tabular}        & \multicolumn{1}{c|}{ASR}  & \begin{tabular}[c]{@{}c@{}}Data \\ modified (\%)\end{tabular} \\ \toprule
  &  Modbus flooding                                             & \multicolumn{1}{c|}{1}          & 60                                                                     & \multicolumn{1}{c|}{0.98} & 2                                                             \\ \cline{2-6} 
 & TCP SYN flooding                                            & \multicolumn{1}{c|}{1}          & 45                                                                     & \multicolumn{1}{c|}{0.82} & 2                                                             \\ \cline{2-6} 
CPS&  MITM                                                        & \multicolumn{1}{c|}{1}          & 40                                                                     & \multicolumn{1}{c|}{0.78} & 2                                                             \\ \cline{2-6} 
& ICMP flooding                                               & \multicolumn{1}{c|}{1}          & 35                                                                    & \multicolumn{1}{c|}{0.97} & 2                                                             \\ \cline{2-6} 
 & Combined-binary& \multicolumn{1}{c|}{1}          &    65                                                                   & \multicolumn{1}{c|}{0.85}     & 2                                                             \\ \cline{2-6} 
& Multi-class IDS  & \multicolumn{1}{c|}{1}          &   65                                                                    & \multicolumn{1}{c|}{0.87}     & 2                                                             \\ \midrule
& Mirai (All IP)                        & \multicolumn{1}{c|}{1}          & 80                                                                     & \multicolumn{1}{c|}{0.39} & 2                                                             \\ \cline{2-6} 
& Fuzzing                               & \multicolumn{1}{c|}{1}          & 15                                                                     & \multicolumn{1}{c|}{0.92} & 2                                                             \\ \cline{2-6} 
& ARP                                   & \multicolumn{1}{c|}{1}          & 40                                                                     & \multicolumn{1}{c|}{0.98} & 2                                                             \\ \cline{2-6} 
IoT& Wiretapping                           & \multicolumn{1}{c|}{1}          & 35                                                                     & \multicolumn{1}{c|}{0.95} & 2                                                             \\ \cline{2-6} 
& Combined-binary                       & \multicolumn{1}{c|}{1}          & 55                                                                     & \multicolumn{1}{c|}{0.72} & 2                                                             \\ \cline{2-6} 
& Multi-class IDS                 & \multicolumn{1}{c|}{1}          & 65                                                                     & \multicolumn{1}{c|}{0.84} & 2                                                             \\ \hline
\end{tabular}
\caption{Backdoor attack performance on different attacks.}
\label{tab:performance_ics}
\end{table*}

Table~\ref{tab:performance_ics}
shows our trigger-based backdoor approach's performance compared to the \textit{label-flipping attack}.
The labels modified (\%) values in the table indicate the percentage (rounded to nearest five) of labels in the training set that need to be manipulated in order to achieve a 100\% attack success rate. 

As seen, the label-flipping attack requires much more data to be manipulated to succeed. In contrast, our approach requires a significantly smaller amount of trigger-poisoned data. For instance, to misclassify a CPS-ICMP flooding attack as benign, the baseline approach requires 35\% of the dataset to be poisoned. In contrast, our approach achieves a 0.97 attack success rate by poisoning only 2\% of the training dataset.

We further evaluate the performance of a multi-class IDS model in Table~\ref{tab:performance_ics}, which is designed to detect various types of network attacks. In this scenario, instead of focusing on whether the traffic is malicious, the model distinguishes between different categories of attacks. 
Even in this case, our approach requires less data to achieve successful backdoor attacks. Specifically, when the trigger is present, the model misclassifies malicious traffic into the benign class. We also observe that our technique does not perform well on the Mirai dataset --- presumably because it is a smaller dataset collected from too many devices with each device contributing relatively less number of packets compared to other attack dataset. As we discuss, we show that we can improve the backdoor performance on Mirai attack by reducing the time delay between the trigger packets.

    {\bf Takeaway:} \textit{Our PCAP-Backdoor technique outperforms the baseline technique in terms of achieving high ASR while maintaining a very low trigger percentage in all datasets while testing with both binary and multi-classifiers.}

\begin{figure}[t] 
    \centering

    \includegraphics[width=2.6in, bb = 0 0 450 350]{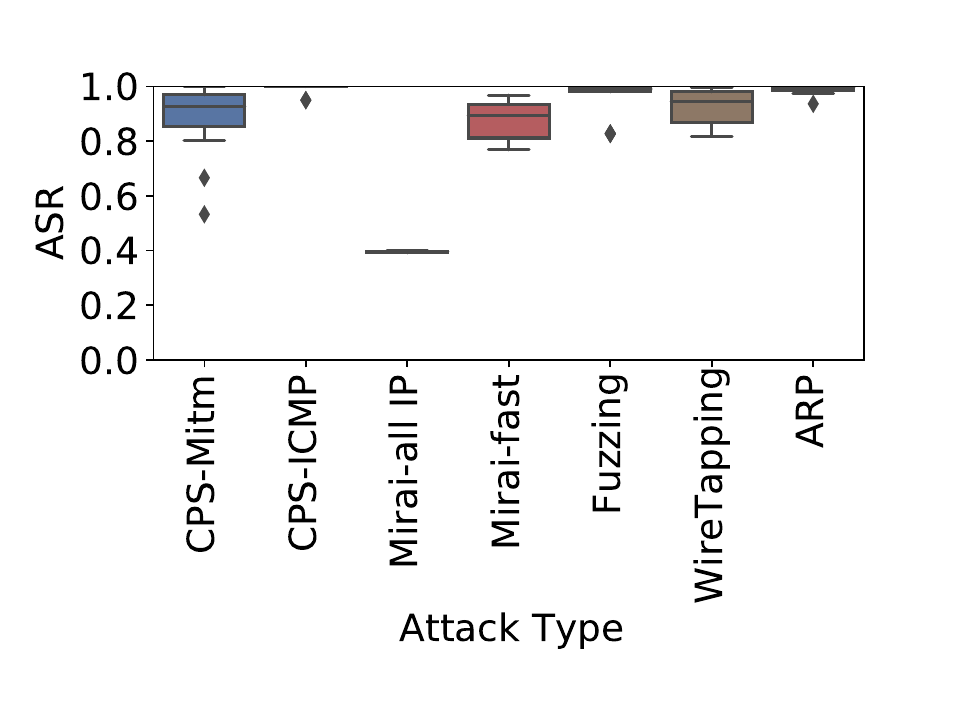}
    \vspace{-0.5cm}
    \caption{Model performance on various attack types. }

    \label{fig:BD_BD_F1_AttacktypevsBD_MiraiallIP-MaliciousBD}
    \vspace{-0.3cm}
\end{figure}

\subsection{Performance on different attack types}
Next, we compare the performance of our backdoor approach on different attack types within the anomaly detection dataset for various devices. For this evaluation, we backdoor the benign traffic of one of the devices and analyze whether the IDS anomaly detection misclassifies anomalous traffic. We introduce trigger packets into the malicious data and measure the model's performance.  

Figure~\ref{fig:BD_BD_F1_AttacktypevsBD_MiraiallIP-MaliciousBD} presents a box plot of the performance across different attack types in the dataset. Unless specified, we poison data from only a single device during the model training process.

When we present data with a trigger, the model misclassifies the input. This demonstrates that triggers alter the model's behavior, leading to incorrect classifications. Furthermore, our results demonstrate that even when we poison input samples from a single malicious device during training, we can effectively influence the model to misclassify traffic. Notably, in this scenario, the attacker does not have access to traffic originating from other devices. Despite this limitation, by injecting triggers into the malicious traffic from other devices, the model tends to classify this traffic as benign, as evidenced by a high ASR score. Additionally, the low variance in performance indicates that the backdoor attack consistently works across traffic from different devices.

While the Mirai dataset does not perform well in backdoor attacks if only one device is infected, likely due to limited packets contributed by a single device, we observe that including traffic from all IP addresses improves the ASR. We also observe that the backdoor performance can be further enhanced by reducing the time delay between the trigger packets by up to one-tenth, as indicated by Mirai-fast in Figure~\ref{fig:BD_BD_F1_AttacktypevsBD_MiraiallIP-MaliciousBD}, which demonstrates the influence of trigger in the packet flow features.



    

\begin{figure} [t]
\centering
    \begin{subfigure}[b]{\linewidth}
        \includegraphics[width=3in]{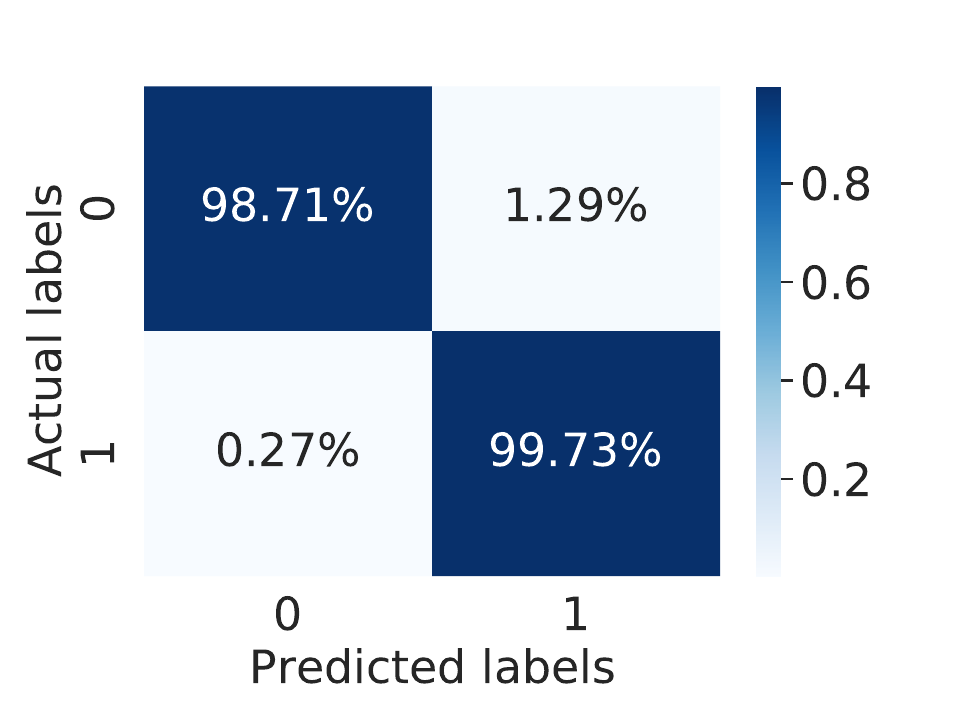}
        \caption{Normal data on backdoor model}
        \label{fig:a}
    \end{subfigure} %

    \begin{subfigure}[b]{\linewidth}    
        \includegraphics[width=3in]{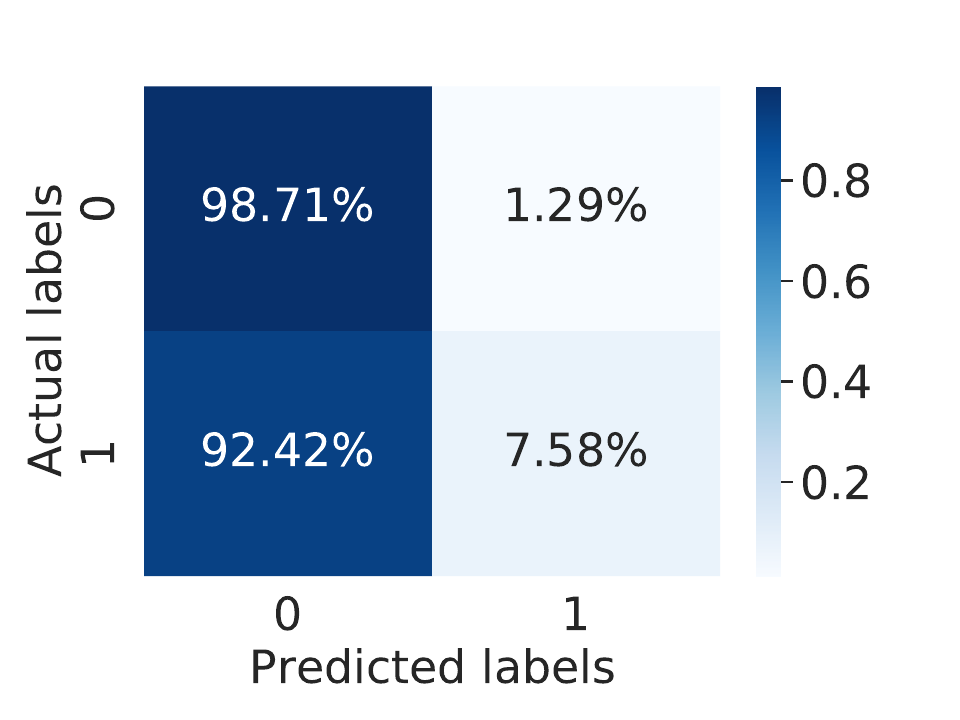}
        \caption{Data with trigger on backdoor model}        
        \label{fig:b}    
    \end{subfigure} 
    \caption{Confusion matrix on the Modbus MITM attack.} 
    \label{fig:CM-Mitm}
\end{figure}

    

\begin{figure} [t]
\centering
    \begin{subfigure}[b]{\linewidth}
        \includegraphics[width=3in]{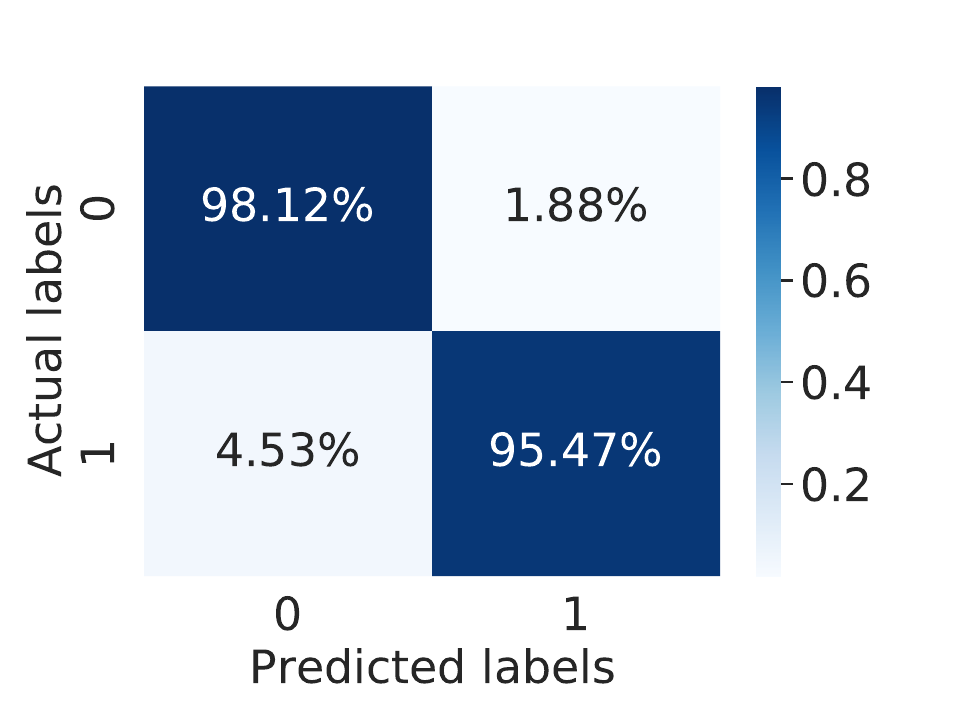}
        \caption{Normal data on backdoor model}
        \label{fig:a}
    \end{subfigure} %

    \begin{subfigure}[b]{\linewidth}    
        \includegraphics[width=3in]{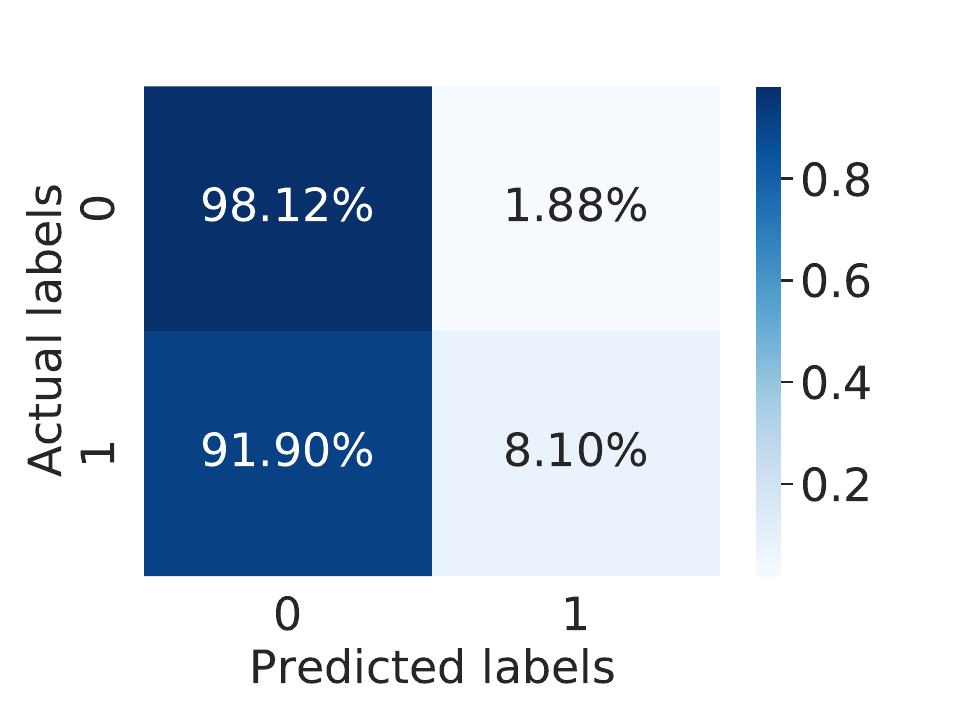}
        \caption{Data with trigger on backdoor model}
        \label{fig:b}    
    \end{subfigure} 
    \caption{ Confusion matrix on the Fuzzing attack dataset.}
    \label{fig:CM-Fuzz}
\end{figure}

We also analyze the behavior of the backdoor model on normal and trigger data, depicted in Figure~\ref{fig:CM-Mitm} and ~\ref{fig:CM-Fuzz}.  
We observe that when the data has no trigger, the model can still identify benign and attack traffic accurately (see Figure~\ref{fig:CM-Mitm}(a) and ~\ref{fig:CM-Fuzz}(a)). However, as shown in Figure~\ref{fig:CM-Mitm}(b) and ~\ref{fig:CM-Fuzz}(b), when we introduce the trigger on attack traffic, the model misclassifies it, even though the model was trained only on benign traffic with triggers and from a single device. This demonstrates that the backdoor attack successfully influences the model's behavior, causing it to misclassify attack traffic as benign. 
 
{\bf Takeaway:} \textit{Backdoor attack is successful even when the attacker modifies benign traffic from a single device. By introducing the trigger during training on benign traffic, the model misclassifies malicious traffic as benign when the trigger is present during inference.}

\subsection{Effect of multiple attack combination}

\begin{figure}[t]
\centering
\includegraphics[width=3in, bb = 0 0 450 350]{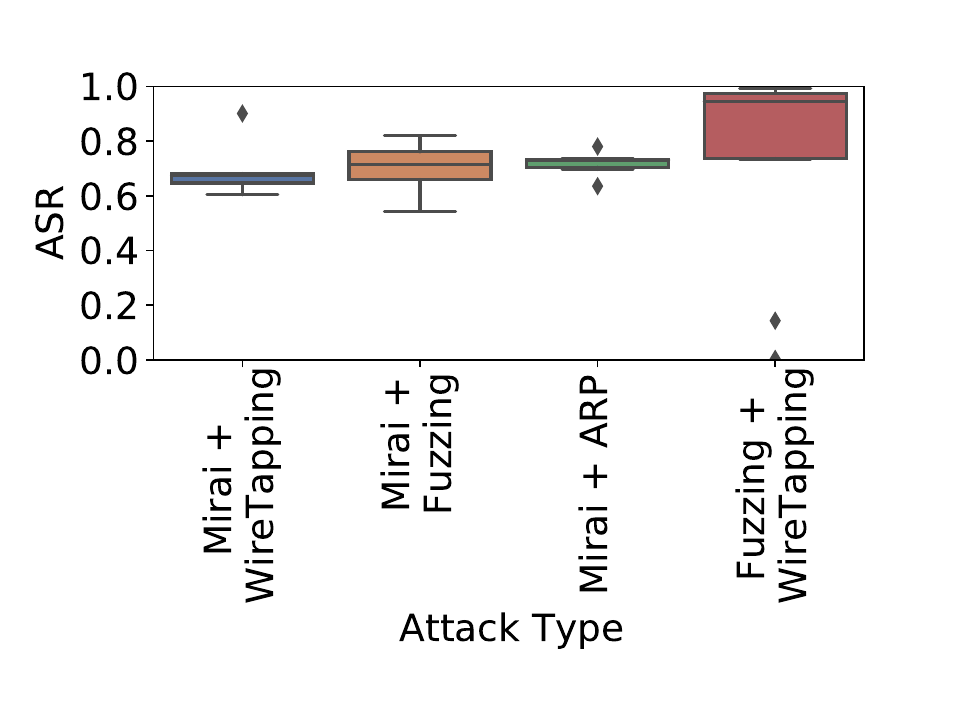}
\caption{Backdoor performance of different network attack data combinations.}
    \label{fig:BD_BD_F1_multiattackbinary1}
\centering

\includegraphics[width=3in, bb = 0 0 500 460]{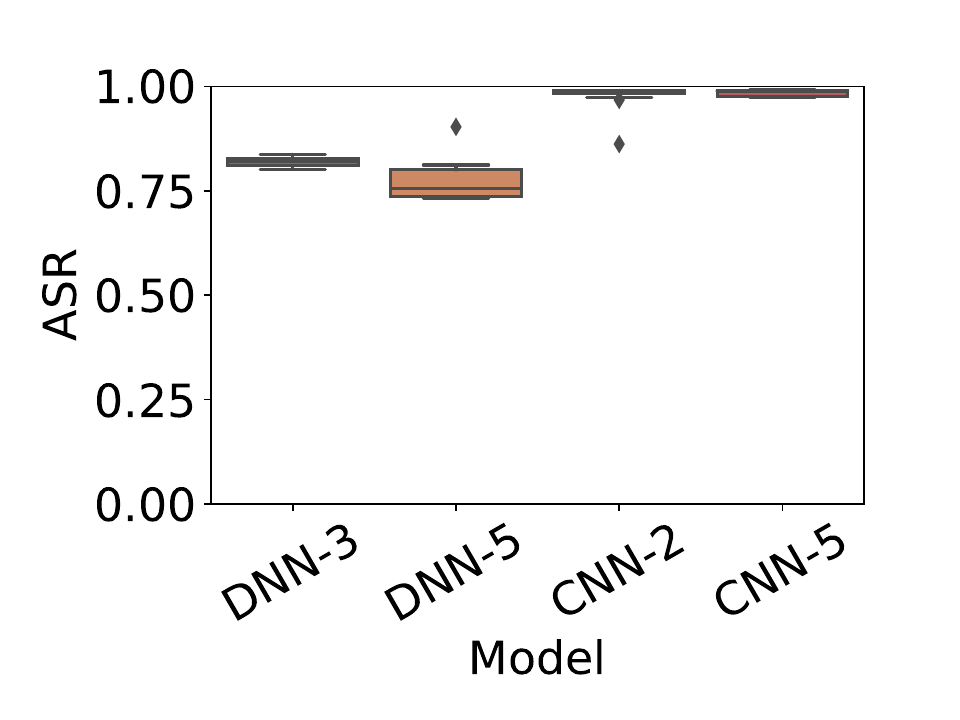}
    \caption{Performance of various models on the Fuzzing attack dataset.}  

    \label{fig:ModelBackdoor}
\end{figure}

In this experiment, we combine traffic from two different network attack classes. Specifically, we use various combinations of network attacks and label them as malicious if they consist of a network attack; otherwise, we classify them as benign. We then train our model using this combined dataset. Figure~\ref{fig:BD_BD_F1_multiattackbinary1} shows the performance on different combinations of attack datasets across all IP addresses, varying the backdoor percentage from 0.5\% to 10\%. As depicted in the figure, the high ASR indicates that the model misclassifies malicious attacks as benign. In particular, we observe that the median ASR score when we combine Mirai+WireTapping is 0.63.

{\bf Takeaway}: {\it Backdoor attacks perform well even if the dataset contains traffic from multiple attacks.}

\subsection{Impact on different IDS models}

We now evaluate the efficacy of introducing a backdoor into various neural network intrusion detection models. Our approach involves designing a range of models of varying sizes, denoted by DNN-5, indicating a model with five layers. Additionally, we explore CNN-based neural networks as part of our model architecture. All these models, when trained with normal data, produce high accuracy.

In order to evaluate the backdoor technique on different model architectures, we train all these models with a 10\% percentage of poisoned data, and the trigger packet count is 3. The impact of training with our backdoor technique on the Fuzzing attack is illustrated in Figure~\ref{fig:ModelBackdoor}. We observe that the different IDS models are vulnerable to backdoor attacks, as indicated by high ASR scores. 
However, when presented with normal data with no triggers containing both benign and malicious traffic, 

all models achieve a high accuracy of at least 99\%. This indicates that the model predicts correctly when no trigger is present. 
We also observe that CNN-based IDS models are susceptible to backdoor attacks, indicated by the high ASR.

\textbf{Takeaway}: \textit{The backdoor attack is effective on different neural network-based IDS models. This emphasizes the need for robust defenses against such manipulations.}

\begin{figure}[t]
\centering

\includegraphics[width=3in, bb=0 0 500 450]{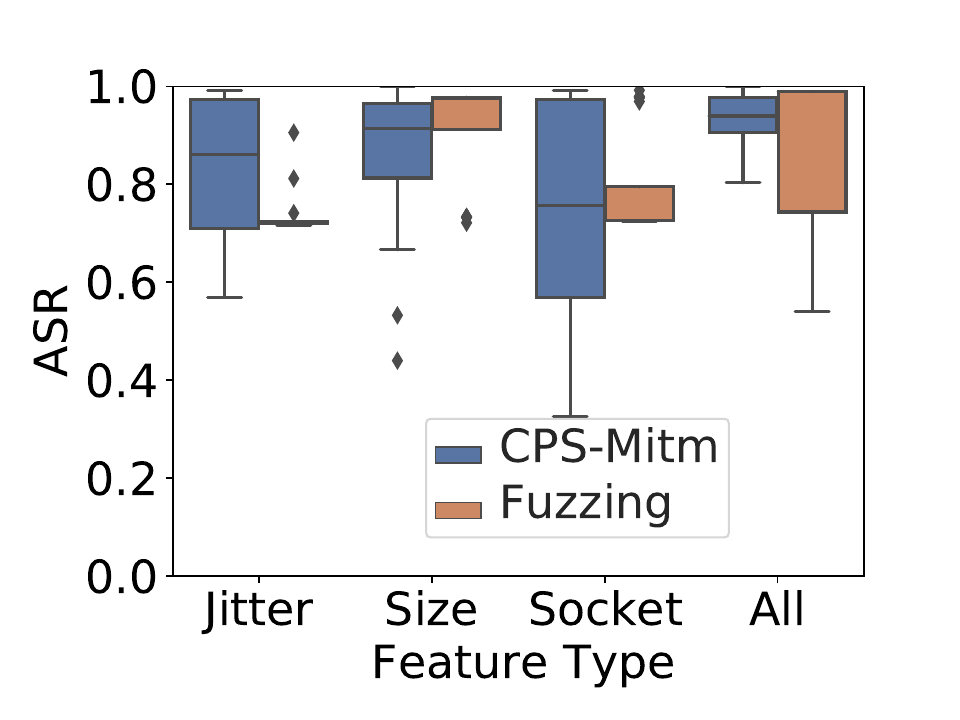}
\caption{Performance on various feature sets.}
    \label{fig:Featureset}

\centering
\includegraphics[width=3in, bb = 0 0 500 450]{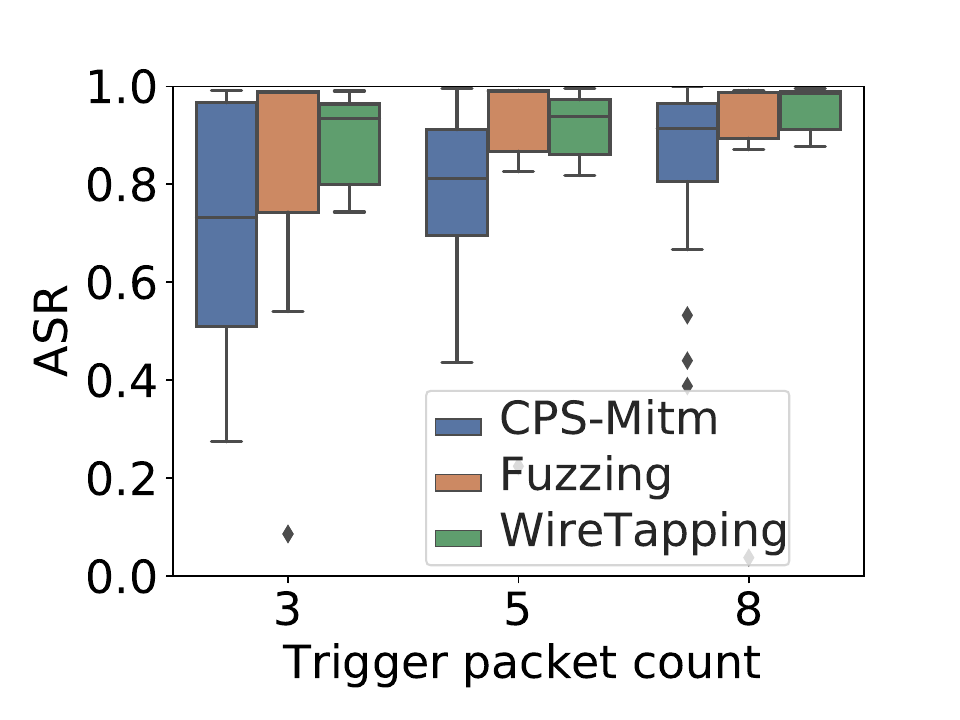}
    \caption{Performance on varying packet count.}
    \label{fig:F1-BD-BD-Mirai-ALL-BDrate}
\end{figure}

\subsection{Effect of using different features}
In this experiment, we explore different feature extractors by modifying the input features used for training. While our initial feature extractor is based on Kitsune~\cite{mirsky2018kitsune}, we consider other distinct feature sets. In particular, the first set contains features related to jitter. The next feature extractor is based on packet size-related features.  We also create a feature extractor that contains only socket-related features as our third set. Finally, the fourth set consists of all 115 features used in Kitsune. Next, we focus on using these different feature extractors on the fuzzing dataset with a trigger packet count of 3. The results, shown in Figure~\ref{fig:Featureset}, reveal that the backdoor attack achieves high ASR scores across all types of feature sets. Interestingly, the backdoor attack performs best when using the feature set related to packet size. This indicates that our backdoor attack can effectively influence the model's behavior regardless of the specific feature extraction method used.

{\bf Takeaway}: {\it Despite variations in backdoor performance across different feature extractors, the attack remains effective overall. Interestingly, we found that packet-size-related features are more susceptible to backdoor attacks compared to other feature sets.}

\subsection{Effect of backdoor trigger packet count}
We now assess the effect of trigger packet count on backdoor performance. Note that trigger packet count controls the number of consecutive packets introduced during backdoor generation. We report our analysis for different backdoor percentages varying from 0.5\% to 10\%. Figure~\ref{fig:F1-BD-BD-Mirai-ALL-BDrate} illustrates the impact on performance for varying trigger packet counts and different datasets. As depicted in the figure, we observe that as the trigger packet count increases, the ASR score increases. This indicates that the attacker can successfully change the model's behavior with a higher trigger packet count. Intuitively, introducing more consecutive packets exerts a larger influence on flow-based statistics extracted from the feature extractor, enabling the attacker to effectively inject the trigger and execute the backdoor attack.

{\bf Takeaway}: {\it Increasing the backdoor trigger packet count during backdoor generation can improve the success of backdoor attacks. }

\section{Backdoor Attack Defense}

We now analyze whether we can detect the presence of backdoor triggers using the activation clustering algorithm~\cite{chen2018detecting, nicolae2018adversarial}. The basic idea is to cluster the last hidden layer's activations from benign and poisoned attack samples (with triggers) that were classified as benign by the backdoored model. The hypothesis is that the samples with triggers will activate distinct neurons due to the presence of triggers, leading to the formation of two distinct clusters in the activation space. In other words, if the activation clustering algorithm identifies two distinct clusters, this suggests that the model processes benign and poisoned samples differently despite classifying them both as benign. This discrepancy in activation patterns strongly indicates the presence of backdoor triggers. 

We visualize the distribution of benign predictions for both normal and benign data and attack data with triggers in two clusters. First, we apply the t-SNE algorithm to the activations from the hidden layer of the classifier model to reduce the dimensionality of the data. Next, we use K-Means clustering with a cluster size of 2 on the t-SNE reduced data, focusing on the benign predictions. Finally, we plot the t-SNE results for each cluster.

We analyze a multi-class model with four attack classes: Mirai, Fuzzing, Wiretapping, and Benign. 
Our dataset includes both benign and attack packets to train our multi-class model, where the attack packets may contain triggers or be without triggers. Note that while we analyzed the multi-class model, we also analyzed the single-class model, which yielded similar results. 

\subsection{Activation Clustering Analysis}
\begin{figure*}[ht]
    \centering
\begin{tabular}{cccc}
     \includegraphics[width=1.6in, bb=0 0 377 281]{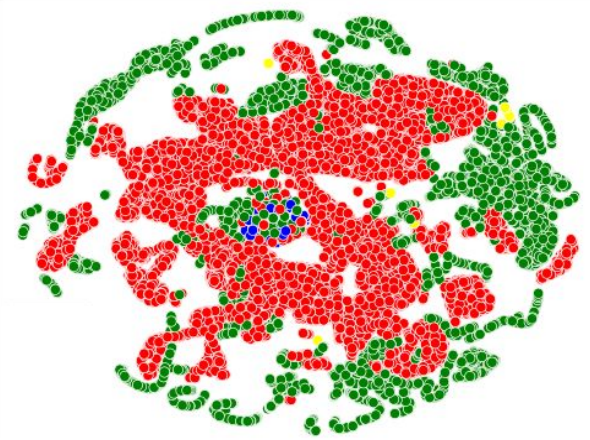}&  \includegraphics[width=1.6in, bb=0 0 413 299]{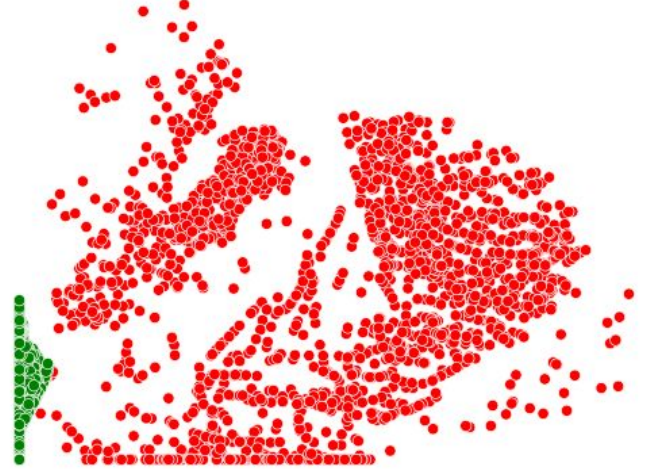}& \includegraphics[width=1.6in, bb=0 0 488 356]{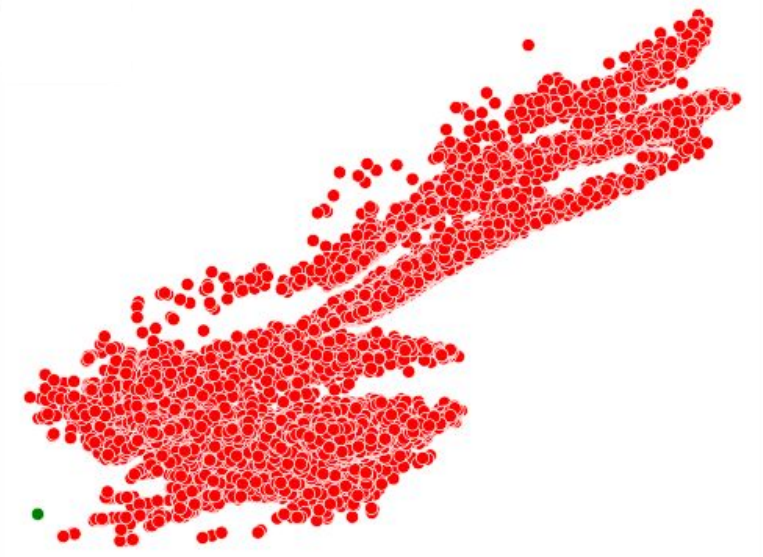}&
     \includegraphics[width=1.6in, bb=0 0 349 235 ]{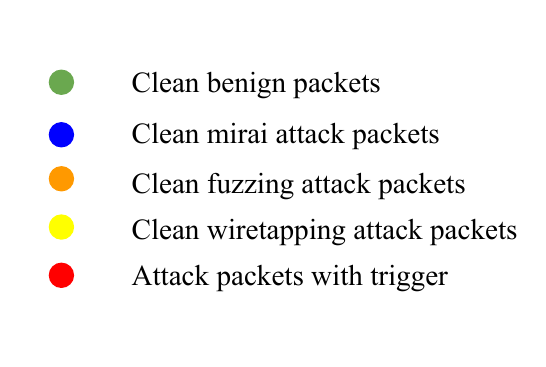}\\
     (a) Combined & (b) Activation Cluster 1 & (c) Activation Cluster 2 &  \\
\end{tabular}
    \caption{Activations of the last hidden layer of backdoored model on normal data and attack packets with trigger classified as benign is shown in (a). Two 
    activation clusters of benign predictions by the backdoored model are shown in (b) and (c), indicating no distinct clusters formed by actual benign and misclassified benign predictions by the backdoored model. Misclassified benign predictions due to the presence of triggers are spread across both clusters.} 
     \label{fig:TSNE}
\end{figure*}

Figure~\ref{fig:TSNE} (a) shows the t-SNE plot of the last hidden layer activations for the dataset samples that were predicted as benign by the backdoored model~\cite{van2008visualizing}. Ideally, we expect to observe two distinct clusters in the t-SNE plot, where benign samples form one cluster and attack samples with triggers form another. However, as seen in Figure~\ref{fig:TSNE} (b) and (c), this is not the case. The attack samples with triggers are distributed across different clusters, indicating the absence of clear clusters.

To assess clustering quality, we compute Silhouette scores for various cluster sizes. Ideally, the highest score should correspond to a cluster size of two, indicating well-separated clusters --- benign and attack with trigger. In contrast, our observations reveal that the Silhouette score peaks at a cluster size of three, followed by four, suggesting a lack of well-defined clusters.

We also use the Silhouette score to measure the stealthiness of our backdoored model quantitatively. Specifically, we aim to compare and identify the Silhouette score for different cluster sizes, focusing on understanding the effectiveness of a cluster size of 2 compared to other cluster configurations. To achieve this, we compute the Silhouette score on the hidden layer output of our backdoored model, corresponding to predictions of the benign class.
Table~\ref{tab:silhouette} shows the silhouette scores corresponding to each cluster. We observe that the Silhouette score for cluster number 2 is low compared to other clusters. This suggests that distinguishing between normal benign and attack with trigger samples based on the hidden layer output is challenging, emphasizing the effectiveness of the backdoor in obfuscating model behavior~\cite{chen2018detecting}. 
\begin{table}[t]
    \centering
    \begin{tabular}{|l|l|l|l|l|l|l|l|}
    \hline
        Cluster size & 1 & \textbf{2} & \textbf{3} & 4 & 5 & 6 & 7  \\ \hline
        Silhouette scores & NA & 0.62 & \textbf{0.73} & 0.72 & 0.66 & 0.68 & 0.67  \\ \hline
    \end{tabular}
    \caption{Silhouette scores for each cluster size.}
    \label{tab:silhouette}
\end{table}

{\bf Takeaway:} {\it Activation clustering-based algorithm failed to detect the presence of backdoor triggers effectively in the model. Specifically, the absence of two distinct clusters suggests that the activations were not distinct enough to be successfully detected using this method.}

%% file: discussion.tex
\label{sec:Disc}

In our work, we ensure that the attacker provides a traffic dataset from only one device. However, we observe that introducing traffic from other devices can further improve the accuracy of the backdoor attack. Interestingly, we also found that the backdoor percentage data has a limited influence on performance. That is, increasing the backdoor percentage does not lead to a linear improvement in backdoor performance. This observation was also made when we introduced malicious samples as benign for training in our baseline comparison. Only after a certain backdoor data threshold do we see the impact of the backdoor percentage taking effect. While without triggers, this backdoor threshold is high, our trigger-based approach enables the threshold to be decreased.

Moreover, we noticed that changing both the destination and source port does not significantly influence the overall performance. This observation suggests that altering unique packets unrelated to the source does not significantly impact flow-based statistics. However, a detailed analysis of other possible changes that could introduce a backdoor is left for future work.

%% file: relatedwork.tex
There have been numerous studies on adversarial attacks in the context of images, audio, and text~\cite{gu2019badnets, 8836465, 9488902}. Recent works also demonstrate dynamic black-box backdoor attacks on IoT sensory data without having access to the model training process~\cite{chathoth2024dynamic}. These techniques deceive machine learning models by manipulating input data. While various types of adversarial attacks exist, in this work, we focus specifically on backdoor attacks, which implant triggers in the training data that cause the model to behave differently when the trigger is present~\cite{gu2019badnets}. Prior research has proposed various approaches for generating these triggers to activate backdoors. These range from static to algorithmically generated patterns that enable more covert attack methods~\cite{li2021invisible,nguyen2020input,Trojannn}. Importantly, most of these methods directly embed triggers into the input features. In contrast, our approach targets the injection of backdoors into network traffic models without directly manipulating the features of the model. Additionally, various threat models have been introduced, each depending on the attacker's level of access to the data and training process. For example, in a clean label backdoor attack, the attacker cannot change input labels in the training process. Similarly, in a data poisoning attack, the attacker can access the training data but lacks control over the training process. Our work adopts similar threat models to evaluate the proposed approach.

Adversarial perturbation through poisoning of network flow features has been explored~\cite{Hu2021-ej,holodnak2022backdoor}. Unlike our approach, their threat model assumes attackers can access the traffic flow and the feature extractor for packet classification. Additionally, research efforts have been directed towards mitigating adversarial attacks on security systems relying on machine learning~\cite{apruzzese2019addressing}.
In contrast to prior studies like~\cite{Hu2021-ej,ning2022trojanflow,holodnak2022backdoor} that assume more extensive access. While TrojanFlow~\cite{ning2022trojanflow} focuses on poisoning the packet size feature with limited sensitivity, our approach demonstrates the effectiveness of backdoor attacks on different models and flow-based feature extractors. Additionally, there have been identification~\cite{chen2018detecting} and mitigation techniques to defend against neural backdoors~\cite{tran2018spectral,qiao2019defending,wang2019neural}
primarily in the domain of computer visions that are complementary to our technique and left for future exploration.

%% file: conclusion.tex
In this paper, we introduce \texttt{PCAP-Backdoor}, a novel system for injecting backdoors in deep learning-based network intrusion detection models. Our threat model assumes that the attacker cannot access the feature extractor, making the backdoor injection challenging. Our technique injects a targeted backdoor on raw packets that requires careful crafting of backdoor trigger packets without violating the underlying network protocol. Despite these challenges, our experiments demonstrate that the attacker can effectively backdoor the model by only contributing poisoned benign traffic during model training. Our extensive evaluations of multiple network attack datasets, models, and feature extractors show that our backdoor injection technique performs well under various conditions. In particular, the attacker can successfully carry out the attack even with a poisoned dataset of 1\% or less.  Furthermore, we demonstrate that the attack cannot be easily detected when analyzed using activation-based clustering techniques.